\documentclass{article}
\usepackage{ijcai23}

\usepackage{times}
\usepackage{soul}
\usepackage{url}

\usepackage{breakurl}
\usepackage[breaklinks]{hyperref}
\usepackage[utf8]{inputenc}
\usepackage[small]{caption}
\usepackage{multirow}
\usepackage{caption}
\usepackage{subfigure}
\usepackage{graphicx}
\usepackage{xspace}
\usepackage{nicematrix}
\graphicspath{{figures/}}
\usepackage{mathtools,commath,amsmath,amssymb,amsfonts,bm}\interdisplaylinepenalty=2500
\usepackage{amsthm}
\usepackage{algpseudocode,algorithm}
\usepackage{tabularx,adjustbox}
\newcolumntype{L}[1]{>{\hsize=#1\hsize\raggedright\arraybackslash}X}%
\newcolumntype{R}[1]{>{\hsize=#1\hsize\raggedleft\arraybackslash}X}%
\newcolumntype{C}[1]{>{\hsize=#1\hsize\centering\arraybackslash}X}%
\usepackage{todonotes}
\usepackage{breqn}
\clearpage
\extrafloats{100}
\RequirePackage{tcolorbox}
\tcbuselibrary{breakable}
\tcbset{parbox=false, left=1ex, right=1ex}

\newcommand{\etal}{\textit{et~al.}\xspace}

\algnewcommand\algorithmicinput{\textbf{Input:}}
\algnewcommand\Input{\item[\algorithmicinput]}
\algnewcommand\algorithmicoutput{\textbf{Output:}}
\algnewcommand\Output{\item[\algorithmicoutput]}
\algnewcommand\algorithmictier{\textbf{Role:}}
\algnewcommand\Tier{\item[\algorithmictier]}

\definecolor{textcolor}{rgb}{0. , 0.5 , 0.9}
\definecolor{imagecolor}{rgb}{1 , 0.7 , 0}
\definecolor{videocolor}{rgb}{0.9 , 0.3 , 0.4}
\definecolor{othercolor}{rgb}{0.8 , 0.4 , 0.9}
\definecolor{attconvcolor}{rgb}{0. , 0.6 , 0.1}
\definecolor{wihtecolor}{rgb}{1 , 1. , 1.}
\definecolor{blackcolor}{rgb}{0, 0. , 0.}
\definecolor{RowColor}{rgb}{0.97, 0.97, 1}

\newcommand{\textshape}{\raisebox{0.5pt}{\tikz\fill[textcolor] (0,0) circle (.8ex);}}
\newcommand{\imageshape}{\raisebox{0.5pt}{\tikz\fill[imagecolor] (0,0) circle (.8ex);}}
\newcommand{\videoshape}{\raisebox{0.5pt}{\tikz\fill[videocolor] (0,0) circle (.8ex);}}

\title{A Pathway Towards Responsible AI Generated Content}
\author{
Chen Chen$^1$
\and
Jie Fu$^2$
\and
Lingjuan Lyu$^1$\thanks{Corresponding author.}
\affiliations
$^1$Sony AI \quad
$^2$BAAI\\
\emails
\texttt{\{ChenA.Chen,Lingjuan.Lv\}@sony.com}\\
\texttt{fujie@baai.ac.cn}
}

\begin{document}
\maketitle

\begin{sloppypar}
\begin{abstract}
AI Generated Content (AIGC) has received tremendous attention within the past few years, with content generated in the format of image, text, audio, video, etc. Meanwhile, AIGC has become a double-edged sword and recently received much criticism regarding its responsible usage. In this article, we focus on 8 main concerns that may hinder the healthy development and deployment of AIGC in practice, including risks from (1) privacy; (2) bias, toxicity, misinformation; (3) intellectual property (IP); (4) robustness; (5) open source and explanation; (6) technology abuse; (7) consent, credit, and compensation; (8) environment. Additionally, we provide insights into the promising directions for tackling these risks while constructing generative models, enabling AIGC to be used more responsibly to truly benefit society.
\end{abstract}

\section{Introduction}
\label{sec:introduction}
\footnotetext{This work is still in progress.}
\textbf{Foundation models for generative AI}.
The success of high-quality AI Generated Content (AIGC) is strongly correlated with the emergence and rapid advancement of large foundation models. 
These models, with their vast capacity, enable the rapid development of domain-specific models, which are commonly employed for the production of various types of content, including images, texts, audio, video, etc. For instance, many text generators are built on the Generative Pre-trained Transformer (GPT)~\cite{radford2018improving} or its derivatives, such as GPT-2~\cite{radford2019language}, GPT-3~\cite{brown2020language}, GPT-3.5, GPT-4, etc. Similarly, numerous text-to-image generators rely on vision-language models such as CLIP~\cite{radford2021learning}, OpenCLIP~\cite{wortsman2022robust}, etc.

\begin{figure}[!htp]
\includegraphics[width=1\columnwidth]{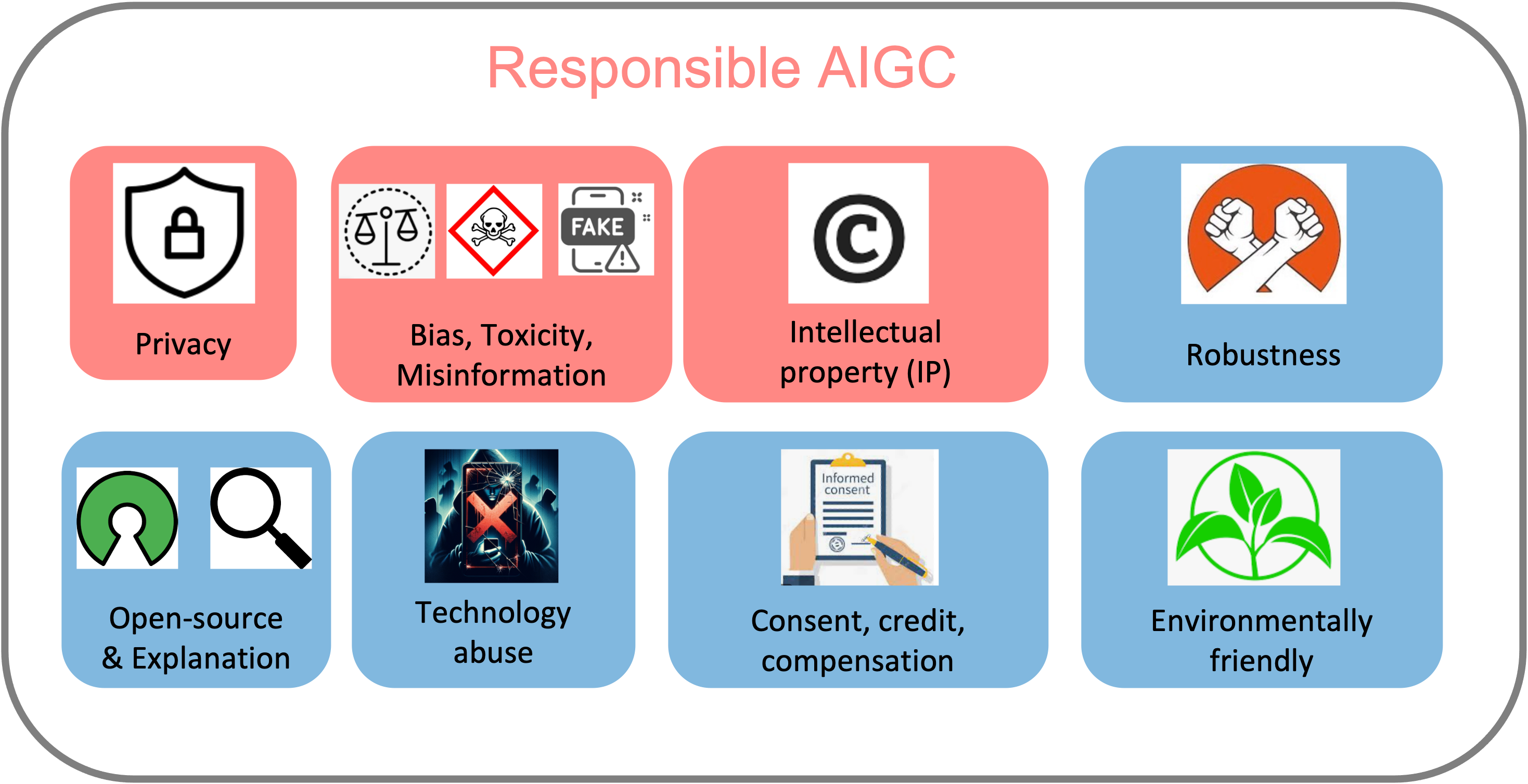}
\caption{The scope of responsible AIGC. Note that some icons are from Shutterstock.}
\label{fig:Responsible_AIGC}
\end{figure}

\paragraph{AIGC applications.}
In recent years, generative modeling has made rapid advances and tremendous progress. 
OpenAI's DALL·E~\cite{ramesh2021zero} was one of the first text-to-image models that had captured widespread public attention. 
It is trained to generate digital images from text descriptions, referred to as ``prompts'', using a dataset of text–image pairs~\cite{brown2020language}. 
Its successor, DALL·E 2~\cite{ramesh2022hierarchical}, which can generate more complex and realistic images, was unveiled in April 2022, followed by Stable Diffusion~\cite{rombach2022high}, which was publicly released in August 2022.
Google, as a rival to OpenAI, presented two text-to-image models that can generate photorealistic images: the diffusion-based model Imagen~\cite{saharia2022photorealistic}, and the Pathways Autoregressive Text-to-Image model (Parti)~\cite{yu2022scaling}.
In addition to text-to-image tasks, diffusion models had been widely used for image-to-image~\cite{saharia2022image,whang2022deblurring} and text-to-video models, such as Runway~\cite{Runway-url}, Make-A-Video~\cite{singer2022make}, Imagen Video~\cite{ho2022imagen}, and Phenaki~\cite{villegas2022phenaki}. 
Stable Diffusion has been adapted for various applications, from medical imaging~\cite{chambon2022adapting} to music generation~\cite{agostinelli2023musiclm}.

In addition to image and video generation, text generation is a popular generative domain. OpenAI's GPT-3~\cite{brown2020language} is a notable example of a large language model (LLM). With a simple text prompt, GPT-3 can produce a piece of writing or an entire essay. It can also assist programmers in writing code. OpenAI has further developed GPT-3.5, an improved version which is better at generating complex text and poetry. 
In 2022, OpenAI launched ChatGPT~\cite{chatgpt-url}, a 175 billion parameter natural language processing (NLP) model that can produce responses in a conversational style. This model combines two popular AI topics: chatbots and GPT-3.5. ChatGPT is a specific chatbot use case wherein the chatbot interacts with a GPT information source. The most recent version of ChatGPT integrated GPT-4~\cite{gpt4} -- OpenAI's most advanced system, which can produce safer and more useful responses.

\begin{figure*}[!htp]
\centering 
\includegraphics[width=2\columnwidth]{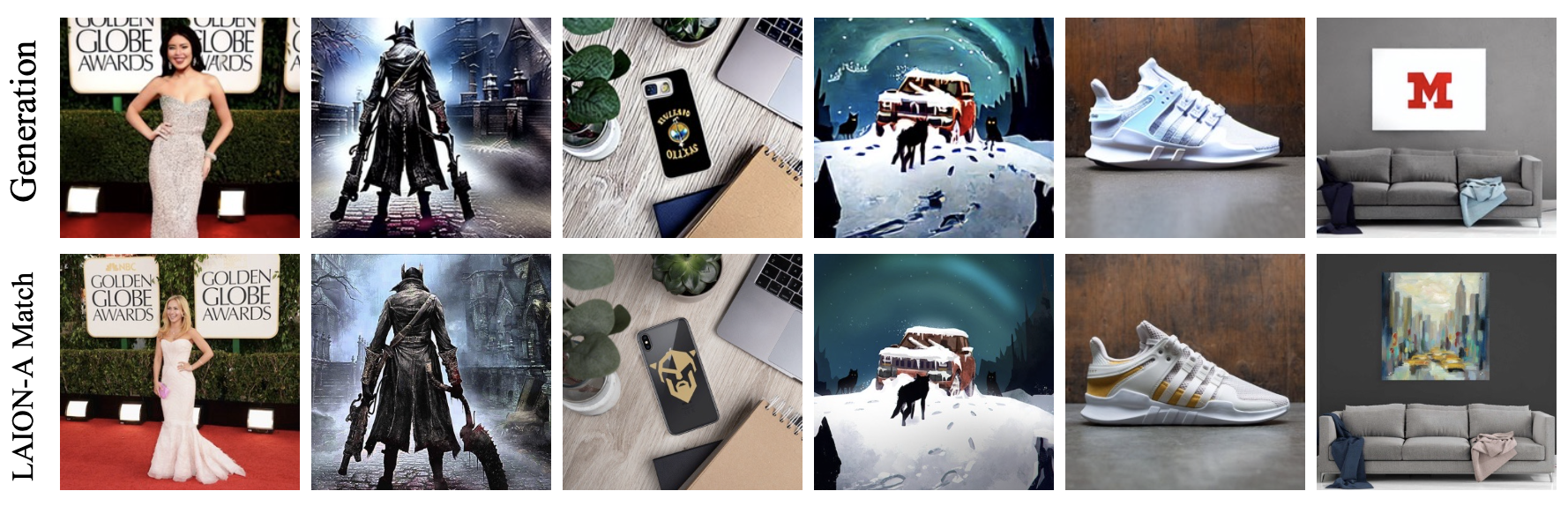}
\caption{A comparison between training images and generated images (by Stable Diffusion).
\textbf{Top row}: generated images. 
\textbf{Bottom row}: closest matches in the training dataset (LAION). 
The comparison shows that Stable Diffusion is able to replicate training data by combining foreground and background objects. 
Image source:~\protect\cite{somepalli2022diffusion}.}
\label{fig:AIGC_privacy}
\end{figure*}

\paragraph{AIGC dispute.}
Despite its popularity, AIGC has raised a range of concerns such as privacy, bias, toxicity, misinformation, intellectual property (IP), and potential misuse of technology. The recent release of ChatGPT has sparked much conversation surrounding its capabilities and potential risks, such as its ability to debug code or compose essays for students~\cite{chatgpt-news-url}.
It is important to consider whether AIGC can be counted as unique creative works or simply replicate content from their training sets. Ideally, AIGC should produce original and distinct outputs, but the source and IP rights of the training data are often unknown due to the use of uncurated web-scale data~\cite{somepalli2022diffusion}.
Furthermore, the powerful memorization of large AIGC models~\cite{carlini2022quantifying,carlini2021extracting} poses a risk of reproducing data directly from the training data~\cite{stable-diffusion-litigation-url}, 
which potentially violates privacy rights and raises legal concerns around copyright infringement and ownership. In addition to the aforementioned privacy and IP issues, as most AIGC models rely on text encoders that are trained on large amounts of data from the internet, hence these learned models may inherent social biases, toxicity, and produce misinformation.

\paragraph{Components in responsible AIGC.} The essential components of responsible AIGC are summarized in Figure~\ref{fig:Responsible_AIGC}. 
Table~\ref{tbl:AIGC-models} lists recent AIGC models and their associated issues related to privacy, bias, toxicity, misinformation, and IP, noting which models have taken proactive actions. 

\begin{table*}[!t]\footnotesize 
\caption{A summary of recent AIGC models and associated issues.
We use dots with different colors to indicate different modalities involved in the models: \textcolor{textcolor}{Text}, \textcolor{imagecolor}{Image}, \textcolor{videocolor}{Video}. 
}
\label{tbl:AIGC-models}
\centering
\scalebox{1}{
\begin{tabularx}{\linewidth}{|X|X|X|X|X|X|X|X|X|}
\hline
 Models & Developer(s) & Initial release & Format & Main technique & Release to public by Mar, 2023
 & Privacy & Bias, toxicity, misinformation & IP
\tabularnewline
\hline
 \textshape{}\imageshape{} DALL·E, DALL·E 2  & OpenAI &Jan, 2021/ Apr, 2022 &Text-to-image & CLIP, diffusion model & No & Deduplication  & Data filtering and reweighting & --- 
\tabularnewline
\hline
\textshape{}\imageshape{} Craiyon (DALL·E Mini)  
& Boris Dayma et al. &Jul, 2021 &Text-to-image & CLIP, diffusion model & No & Deduplication & --- & --- 
\tabularnewline
\hline
\textshape{}\imageshape{} Stable Diffusion 
& CompVis; Runway; Stability AI &Aug, 2022 &Text-to-image  & CLIP, diffusion model & Yes
&--- & Data filtering & --- 
\tabularnewline
\hline
\textshape{} ChatGPT & OpenAI &Dec, 2022 & Text-to-text & GPT-3.5, reinforcement learning & No & Refusing to provide private information (e.g., phone number) & Data filtering, building tools to screen harmful model outputs, etc. 
& Classifier 
\tabularnewline
\hline
\textshape{}\imageshape{} Point-E & OpenAI &Dec, 2022 &Text-to-3D model & GLIDE, diffusion model & No & --- & --- & --- 
\tabularnewline
\hline
\textshape{}\imageshape{} Midjourney's algorithm & Midjourney &Mar, 2022 &Text-to-image & Unknown & No & --- & --- & DMCA takedown policy 
\tabularnewline
\hline
\textshape{}\imageshape{} Imagen 
& Google Brain &Dec, 2022 &Text-to-image & BERT, T5, CLIP, diffusion model & No & --- & Data filtering & --- 
\tabularnewline
\hline
\textshape{}\imageshape{} Parti 
& Google Brain &Dec, 2022 &Text-to-image & ViT-VQGAN, autoregressive model & No & --- & Prompt filtering, output filtering, and model recalibration & Adding watermark 
\tabularnewline
\hline
\textshape{}\imageshape{}\videoshape{} Video diffusion, 
Imagen Video & Google Brain &Dec, 2022 &Text-to-video & Diffusion model & No & --- & Prompt filtering and output filtering & --- 
\tabularnewline
\hline
\textshape{}\imageshape{}\videoshape{} Make-A-Video & Meta &Dec, 2022 &Text-to-video & CLIP, Pseudo-3D convolutions, diffusion model & No & --- & Data filtering & Adding watermark 
\tabularnewline
\hline
\textshape{}\imageshape{} CogView, CogView 2 & Tsinghua University, Alibaba, BAAI & May, 2021 & Text-to-image & VQVAE, autoregressive model & No & --- & --- & ---
\tabularnewline
\hline
\textshape{}\imageshape{}\videoshape{} CogVideo & Tsinghua University, BAAI & May, 2022 & Text-to-video & CogView 2 & No & --- & --- & --- 
\tabularnewline 
\hline
\end{tabularx}}
\end{table*}

\section{Privacy}

\subsection{Privacy leakage in foundation models} 

Large foundation models are known to be vulnerable to privacy risks, and it is possible that AIGC models that build upon these models could also be subject to privacy leakage. Previous research has demonstrated that large language models such as GPT-2 can be vulnerable to privacy attacks, as attackers can generate sequences from the trained model and identify those memorized from the training set~\cite{carlini2021extracting}.
Kandpal \etal~\cite{kandpal2022deduplicating} have attributed the success of these privacy attacks to the presence of duplicated data in commonly used web-scraped training sets. It has been demonstrated that a sequence that appears multiple times in the training data is more likely to be generated than a sequence that occurred only once. This suggests that deduplication could be used as a potential countermeasure in privacy-sensitive applications.

\subsection{Privacy leakage in generative models}
The replication behavior in Generative Adversarial Networks (GANs) has been studied extensively~\cite{meehan2020non,feng2021gans,webster2021person}. Due to the fact that AIGC models are trained on large-scale web-scraped data~\cite{rombach2022high,ramesh2022hierarchical,saharia2022photorealistic}, the issue of overfitting and privacy leakage becomes especially relevant. For instance, the model card of Stable Diffusion recognized that it memorized duplicate images in the training data~\cite{stable-diffusion-model-card-url}. Somepalli \etal~\cite{somepalli2022diffusion} also demonstrated that Stable Diffusion blatantly copies images from its training data, and the generated images are simple combinations of the foreground and background objects of the training dataset.
Moreover, the system occasionally displays the ability to reconstruct memories, producing objects that are semantically equivalent to the original without being identical in pixel form.
The existence of such images raises concerns about data memorization and the ownership of diffusion images. 

\begin{figure}[!htp]
\centering 
\includegraphics[width=0.99\columnwidth]{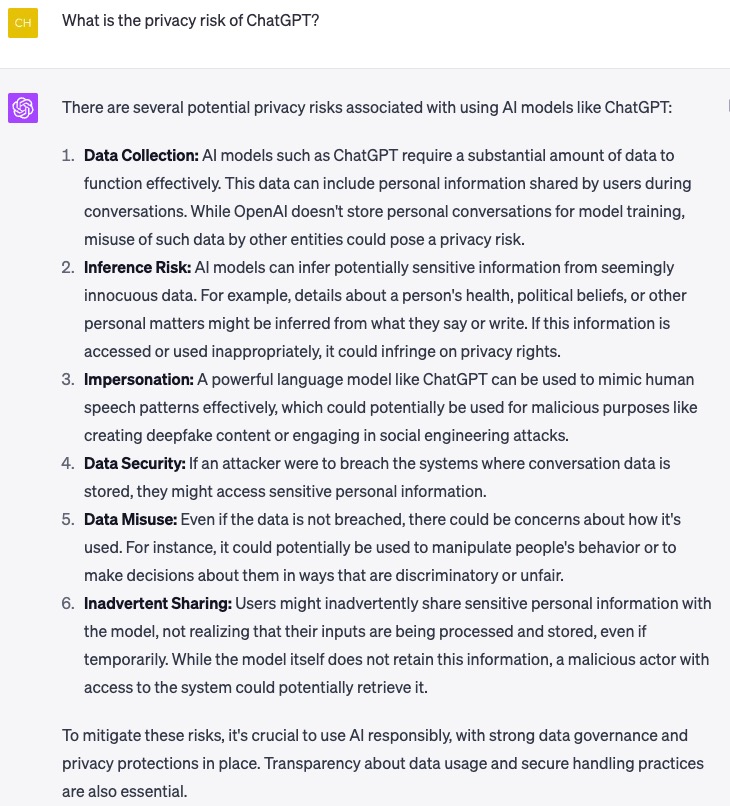}
\caption{An answer to ``What is the privacy risk of ChatGPT" by ChatGPT (GPT-4, May 12, 2023 version).
}
\label{fig:chatgpt}
\end{figure}

Similarly, 
Melissa Heikkilä~\cite{AI-models-spit-url} reported that Google's Imagen can leak photos of real people and copyrighted images.
In Matthew Butterick's recent litigation~\cite{stable-diffusion-litigation-url}, 
he pointed out that because all visual information in the system is derived from copyrighted training images, the images produced are necessarily works derived from those training images, regardless of their outward appearance.
DALL·E 2 also encountered similar problems. It can sometimes reproduce images from its training data rather than creating new ones. OpenAI found that this image regurgitation occurs due to images being replicated many times in the dataset~\cite{dalle2-mitigations-url}. 
Similarly, when we asked ChatGPT "What is the privacy risk of ChatGPT", it responded with multiple potential privacy risks, as illustrated in Figure~\ref{fig:chatgpt}.

\subsection{Privacy actions}
Although a complete resolution to the privacy issues mentioned above has not been achieved, companies and researchers have taken proactive steps to address these issues, such as introducing warning messages and detecting replicated content.

At the industry level, Stability AI has recognized the limitations of Stable Diffusion, such as the potential for memorization of replicated images in the training data. 
To address this, they provide a website~\cite{clip-retrieval-url} to support the identification of such memorized images. 
In addition, art company Spawning AI has created a website called "Have I Been Trained"~\footnote{\url{https://haveibeentrained.com}} 
to assist users in determining whether their photos or works have been used as AI training materials. 
OpenAI has taken steps to address privacy concerns by reducing data duplication through deduplication~\cite{dalle2-mitigations-url}. 
Furthermore, companies such as Microsoft and Amazon have implemented measures to prevent employee breaches of confidentiality by banning the sharing of sensitive data with ChatGPT, given that this information could be utilized for training data for future versions of ChatGPT~\cite{chatgpt-theft-url}. 
At the academic level, researchers~\cite{somepalli2022diffusion} have studied image retrieval frameworks to identify content duplication, while Dockhorn \etal~\cite{dockhorn2022differentially} have proposed differentially private diffusion models to guarantee privacy in generative models. Zhuang \etal~\cite{zhuang2023foundation} proposed to adopt federated learning for the privacy-preserving and responsible development of foundation models.

Existing privacy measures may not be inadequate to meet the demands of privacy. It is essential to explore more reliable detection systems for data replication in generative models, and to further investigate memorization and generalization in current and future AIGC models. Designing more faithful metrics for the privacy assessment on the reconstructed or generated images is also worthwhile to explore~\cite{sun2023privacy}.

\section{Bias, toxicity, misinformation}

\subsection{Problematic datasets}
Since the training data used in AI models are collected in the real world, they can unintentionally reinforce harmful stereotypes, exclude or marginalize certain groups, and contain toxic data sources, which can incite hate or violence and offend individuals~\cite{weidinger2021ethical}. For example, the LAION dataset~\cite{schuhmann2021laion}, which is used to train diffusion models, has been criticized for containing problematic content related to social stereotyping, pornography, racist slurs, and violence.

Although some AIGC models like Imagen~\cite{saharia2022photorealistic} try to filter out undesirable data, such as pornographic imagery and toxic language, the filtered data can still contain sexually explicit or violent content. Moreover, recent research works~\cite{prabhu2020large,birhane2021multimodal} have pointed out that these unfiltered datasets utilized for training frequently encompass social biases, repressive perspectives, and derogatory connections towards underrepresented communities. Google's Imagen Video~\cite{ho2022imagen} is trained on a combination of the LAION-400M image-text dataset and their internal dataset, and Google is concerned that its Imagen tool could be used to generate harmful content. 
Meanwhile, this dataset inherits social biases and stereotypes that are difficult to remove.

\subsection{Problematic AIGC models}
Models trained, learned, or fine-tuned on the aforementioned problematic datasets without mitigation strategies can inherit harmful stereotypes, social biases, and toxicity, leading to unfair discrimination and harm to certain social groups~\cite{weidinger2021ethical}. 

\begin{figure}[!htp]
\centering 
\subfigure{
		\includegraphics[width=0.3\linewidth]{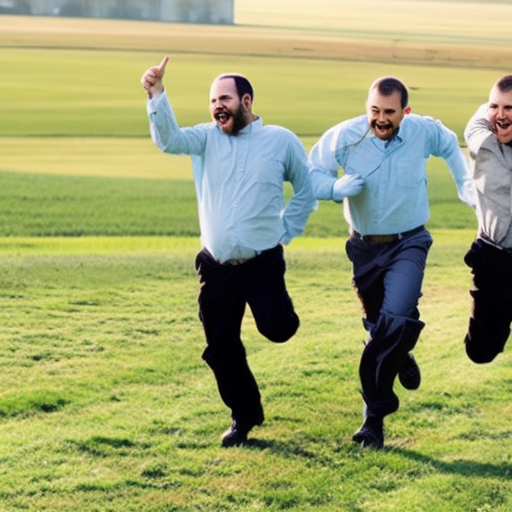}
	}
\subfigure{
		\includegraphics[width=0.3\linewidth]{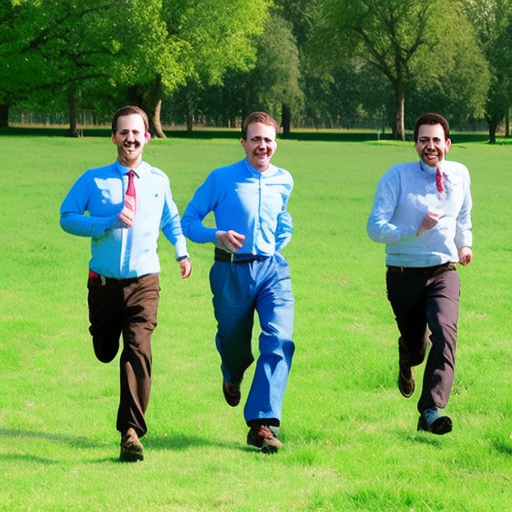}
	}
\subfigure{
		\includegraphics[width=0.3\linewidth]{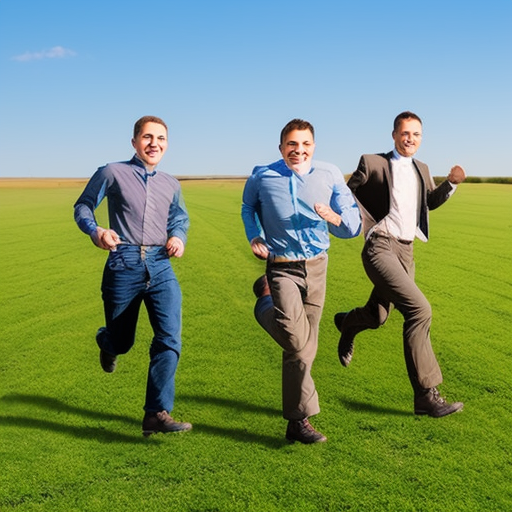}
	}
\subfigure{
		\includegraphics[width=0.3\linewidth]{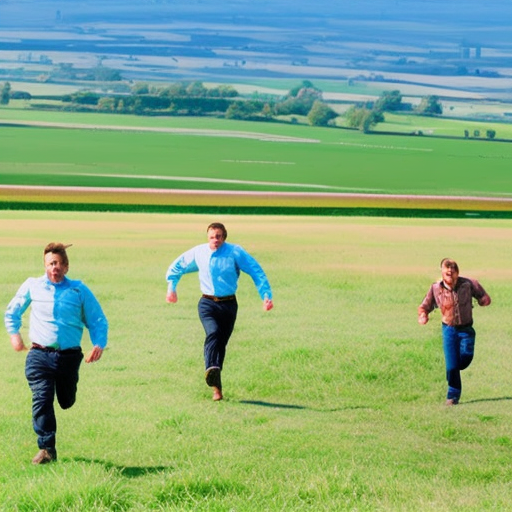}
	}
\subfigure{
		\includegraphics[width=0.3\linewidth]{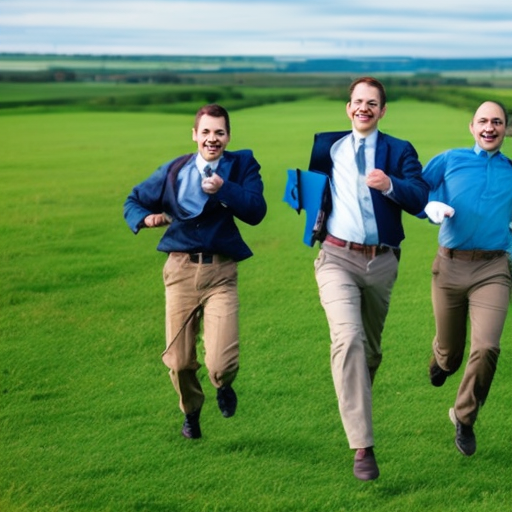}
	}
\subfigure{
		\includegraphics[width=0.3\linewidth]{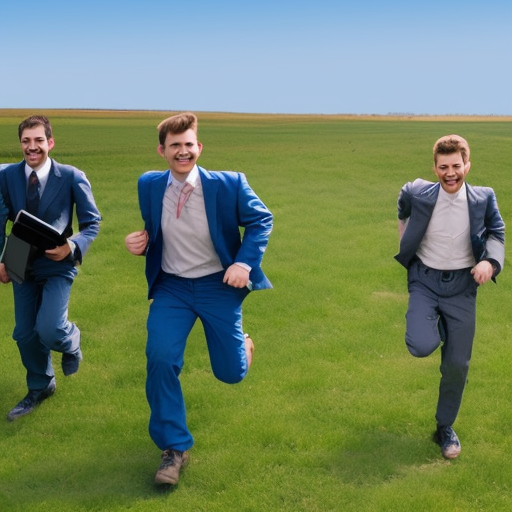}
	}
\subfigure{
		\includegraphics[width=0.3\linewidth]{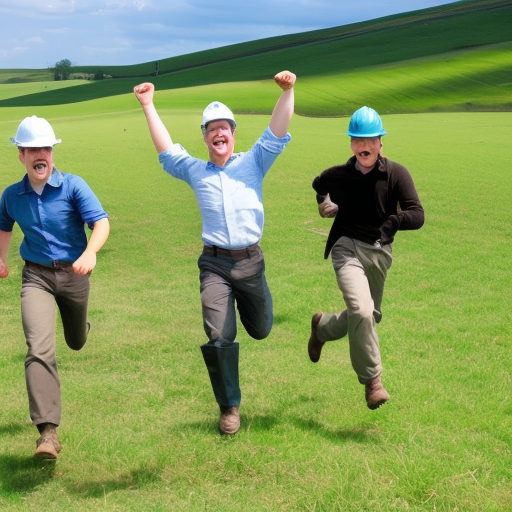}
	}
\subfigure{
		\includegraphics[width=0.3\linewidth]{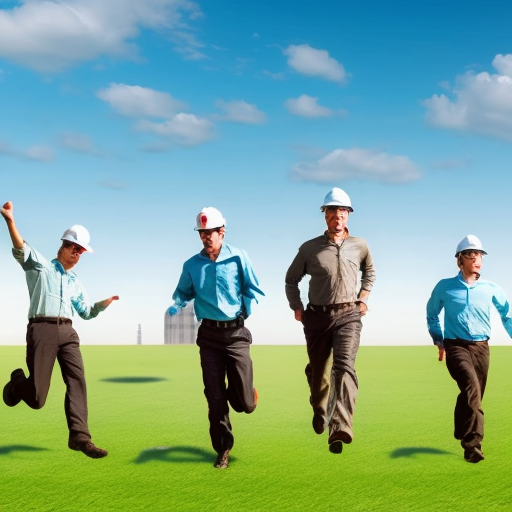}
	}
\subfigure{
		\includegraphics[width=0.3\linewidth]{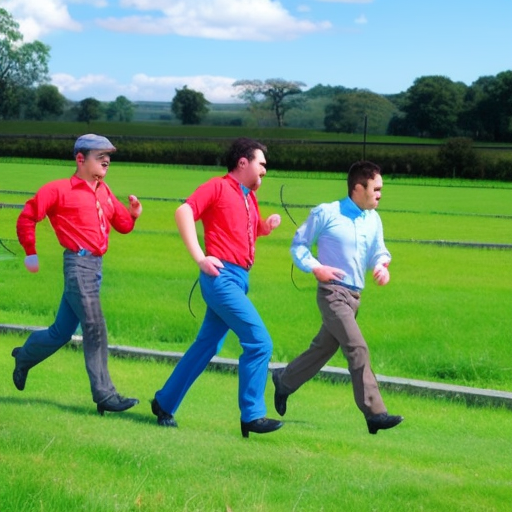}
	}
\caption{Images generated with the text ``Three engineers running on the grassland" by Stable Diffusion v2.1. There are 28 people in the 9 images, all of them are male. Moreover, none of them belong to the neglected racial minorities. This shows a huge bias of Stable Diffusion v2.1.}
\label{fig:unfair}
\end{figure}

For example, Stable Diffusion v1 was trained primarily on the LAION-2B data set, which only contains images with English descriptions~\cite{stable-diffusion-model-card-url}. As a result, the model was biased towards white, western cultures, and prompts in other languages may not be adequately represented. Follow-up versions of the Stable Diffusion model were fine-tuned on the filtered versions of the LAION dataset, but the bias issue still occurs~\cite{stable-diffusion-url}. To illustrate the inherent bias in Stable Diffusion, we tested a toy example on Stable Diffusion v2.1. As shown in Figure~\ref{fig:unfair}, images generated with the prompt ``Three engineers running on the grassland" were all male and none of them belong to the neglected racial minorities, indicating a lack of diversity in the generated images.

Similarly, DALLA·E and DALLA·E 2 
exhibited negative stereotypes against minoritized groups~\cite{dalle2-bias-url}.
Google's Imagen~\cite{saharia2022photorealistic} also encoded several social biases and stereotypes, such as generating images of people with lighter skin tones and aligning with western gender stereotypes. These biases can lead to unfair discrimination and harm to certain social groups. Even when generating non-human images, Imagen has been shown to encode social and cultural biases~\cite{imagen-problematic-url}.
Due to these issues, 
most companies decided not to make their AIGC models available to the public to avoid criticism and potential fine from government.

Beyond above issues, there is also a risk of misinformation when AIGC models provide inaccurate, false even harmful answers or responses~\cite{sun2023pushing,xie2023defending}. For example, ChatGPT and its derivatives are notorious for their hallucination issues, i.e., the generated content 
may appear to be accurate and authoritative, but it could be completely inaccurate. Therefore, it can be used for misleading purposes in schools, laws, medical domains, weather forecasting, and anywhere else. 
For example, the answer on medical dosages that ChatGPT provides could be inaccurate or incomplete, potentially leading to the user taking dangerous or even life-threatening actions~\cite{bickmore2018patient}.
Prompted misinformation on traffic laws could cause accidents and even death if drivers follow the false traffic rules.
ChatGPT also exhibits verbosity and overuse of certain phrases. For instance, it repeatedly states that it is a language model trained by OpenAI. These issues are due to biases inherent in training data, as trainers tend to prefer longer answers that appear more comprehensive~\cite{chatgpt-url}. 

\subsection{Bias, toxicity, misinformation mitigation}
The quality of the content generated by AIGC models is inextricably linked to the quality of the training corpora. One noticeable point is that while problems such as biases, toxicity and stereotypes can be reduced in the source datasets, they can still be propagated even exacerbated during the training and development of AIGC models. For example, although some companies such as Google try to filter out undesirable data before training Imagen~\cite{saharia2022photorealistic}, such as pornographic imagery and toxic language, the filtered data can still contain sexually explicit or violent content. OpenAI took extra measures to ensure that any violent or sexual content was removed from the training data for DALLA·E 2 by carefully filtering the original training dataset.
However, filtering can introduce biases into the training data that can then be propagated to the downstream models. To address this issue, OpenAI developed pre-training techniques to mitigate the consequent filter-induced biases~\cite{dalle2-mitigations-url}. Overall, it is crucial to evaluate the existence of bias and toxicity 
throughout the entire lifecycle of data usage, rather than staying solely at the data source level. Additionally, there is a challenge in defining a truly fair and non-toxic dataset. The extent and nature of these issues within AIGC models have not yet been comprehensively investigated.

In terms of misinformation and hallucination prevention, it is vital to analyze the root reasons behind them. AI hallucinations can occur for several reasons~\cite{AI-hallucinations-url}, including: (1) Insufficient, outdated, or low-quality training data. An AI model is only as good as the data it's trained on. If the AI tool doesn't understand the input prompt or doesn't have sufficient information, it'll rely on the limited dataset it's been trained on to generate a response—even if it's inaccurate; (2) Overfitting. When an AI model is trained on a limited dataset, it may memorize the inputs and appropriate outputs. This leaves it unable to effectively generalize new data, resulting in AI hallucinations; (3) Use of idioms or slang expressions. If a prompt contains an idiom or slang expression that the AI model hasn't been trained on, it may lead to nonsensical outputs; (4) Adversarial attacks. Prompts that are deliberately designed to confuse the AI can cause it to produce AI hallucinations. 

To defend against misinformation and hallucination, Elena~\cite{AI-hallucinations-url} recommended 6 ways to prevent AI hallucinations, including (1) Limit the possible outcomes; (2) Pack in relevant data and sources unique to you; (3) Create a data template for the model to follow; (4) Give the AI a specific role—and tell it not to lie; (5) Tell it what you want—and what you don't want; (6) Experiment with the temperature which controls the randomness of model results. Sun \etal~\cite{sun2023pushing} recently adopted the self-verification strategy to address the hallucination issue of LLMs. 
Xie \etal~\cite{xie2023defending} proposed the psychologically inspired self-reminder technique that can efficiently and effectively mitigate against jailbreaks without further training.

It is also essential to regularly update the training corpora used by AIGC models with the most recent information to ensure that AI-driven models reflect the current state of society, thus reduce misinformation and hallucination. This will help prevent information lag and hallucination, and ensure that the models remain updated, relevant, and beneficial to society.
Lazaridou \etal~\cite{lazaridou2021pitfalls} showed that transformer models cannot accurately predict data that did not fall into training data period. This is because test data and training data come from different periods, and increasing model size does not improve performance. It is thus essential to incorporate new training data and update the model regularly. 
Gathering user feedback is also an 
effective way to keep models updated. Companies such as OpenAI actively seek feedbacks from users to identify harmful outputs that could arise in real-world scenarios, as well as to uncover and mitigate novel risks~\cite{chatgpt-url} in a timely manner. Actually, GPT-4 had incorporated an additional safety reward signal during Reinforcement Learning from Human Feedback (RLHF) training to reduce harmful outputs by training the model to refuse requests for such content~\cite{gpt4}. By involving users in the feedback loop, AIGC developers can better understand the potential consequences of their models and take corrective actions to minimize any negative impacts. 

\section{IP Protection}
As AIGC continues to advance in sophistication and popularity, it raises questions about the origin of content for copyright purposes and whether AI-generated content should be entitled to the same intellectual property protections as content created by humans.

\subsection{Difficulty of IP infringement detection}

\textbf{Traditional understanding of copyright.}
Copyright law generally protects original works of authorship that are created by human authors and are fixed in a tangible form~\cite{copyright-url}. 
For a work to be eligible for copyright protection, it needs to be expressed in a tangible form, either physical or digital, 
such as a book, painting, or computer file.

\textbf{Difficulty of copyright definition in AIGC.}
The ownership and protection of generated content have raised a significant amount of concern and debate. It remains unclear whether such generated content should be considered original works eligible for copyright protection under current laws.

There are many different notions of replication from AIGC. Somepalli \etal~\cite{somepalli2022diffusion} gave an (informal) definition as follows:
\emph{An image is considered to contain replicated content if it includes an object that is identical to an object in a training image, regardless of minor variations in appearance resulting from data augmentation, whether the object is in the foreground or background}.

In fact, addressing AI copyright issues is a complex task that involves several factors, including: (1) unclear regulations on data collection, usage, rights confirmation, and commercial use of data; (2) the need for a fair benefit distribution mechanism for contributors; (3) the lack of a unified legal understanding of AIGC copyright worldwide, with disputes over ownership still unresolved; and (4) difficulties in identifying all original works used to train AIGC models, as these models can generate an unlimited amount of content, making it impossible to test all of it.

\subsection{IP infringement examples}
There is a risk of copyright infringement with the generated content if it copies existing works, whether intentionally or not, raising legal questions about IP infringement.

In November 2022, Matthew Butterick filed a class action lawsuit against Microsoft's subsidiary GitHub, accusing that their product Copilot, a code-generating service, violated copyright law~\cite{github-copilot-investigation-url}. 
The lawsuit centers around Copilot's illegal use of licensed code sections from the internet without attribution. Texas A\&M professor Tim Davis also provided examples of his code being copied verbatim by Copilot~\cite{devs-dont-rely-url}. Although Microsoft and OpenAI have acknowledged that Copilot is trained on open-source software in public GitHub repositories, Microsoft claims that the output of Copilot is merely a series of code ``suggestions" and does not claim any rights in these suggestions. Microsoft also does not make any guarantees regarding the correctness, security, or copyright of the generated code.

In addition to code generation, text-to-image generative models like Stable Diffusion also faced accusations of infringing on the creative work of artists, as they are trained on billions of images from the Internet without the approval of the IP holders, which some argue is a violation of their rights. This is evident in Stable Diffusion, which has generated images with the Getty Images’ watermark on them~\cite{getty-images-url}. Somepalli \etal~\cite{somepalli2022diffusion} also presented evidence suggesting that Stable Diffusion copies from the data on which it was trained on. 
While Stable Diffusion disclaims any ownership of generated images and allows users to use them freely as long as the image content is legal and non-harmful, this freedom raises questions about ownership ethics. 

\subsection{IP problem mitigation}
To mitigate IP concerns, many companies have started implementing measures to accommodate content creators. Midjourney, for instance, has added a DMCA takedown policy to its terms of service, allowing artists to request the removal of their work from the dataset if they suspect copyright infringement~\cite{midjourney-terms-of-service-url}. 
Similarly, Stability AI plans to offer artists the option of excluding themselves from future versions of Stable Diffusion~\cite{stable-diffusion-opt-out-url}. OpenAI has released a classifier that can distinguish between text generated by AI and that written by humans. However, this tool should not be relied exclusively on for critical decisions.

In addition to above attempts, watermarks can be extremely useful in tracking IP violations or detecting the origin of the generated content~\cite{he2022protecting,he2022cater,peng2023you}. Wang \etal ~\cite{wang2023did,wang2023detect} recently utilized novel watermark techniques to conduct origin attribution of AI-generated images and detect the unauthorized data usages in text-to-image diffusion models. In light of the growing popularity of AIGC, the need for watermarking is becoming increasingly pressing. 
OpenAI is developing a watermark to identify text generated by its GPT model. It could be a valuable tool for educators and professors to detect plagiarism in assignments generated with such tools.
Google has already applied a Parti watermark to all images it releases. John Kirchenbauer \etal~\cite{kirchenbauer2023watermark} proposed a watermark to detect whether the text is generated by an AI model. Still, they only tested it on the smaller open-source language model OPT-6.7B from Meta, leaving its performance on the larger and more widely used ChatGPT model unknown.

In general, the emergence of AIGC presents significant IP concerns and challenges that demand immediate attention. It is essential for technologists, lawyers, and policymakers to recognize these issues and work together to ensure that the intellectual property rights of human creators are protected.

\section{Robustness}
Previous studies have demonstrated that large models or foundation models trained on the unlabelled data can be backdoored~\cite{pan2023asset,shejwalkar2023perils}. 
This poisoning effect could cause catastrophic damage to downstream applications that depend on the compromised foundation or generative models. For example, a diffusion model with a hidden ``backdoor" could carry out malicious actions when it encounters a specific trigger pattern during data generation~\cite{chou2022backdoor,zhang2022fine,sun2023defending}. How to sift out clean data for training~\cite{zeng2022sift} matters a lot for model robustness. 

Beyond the poisoning attack during training phase, the recent emergence of jailbreak attacks~\cite{link_jailbreak_chatgpt,link_jailbreak_chat} which use adversarial prompts to bypass the deployed ChatGPT’s ethics safeguards and engender harmful responses notably threatens the responsible and secure use of ChatGPT~\cite{xie2023defending}. 
Xie \etal~\cite{xie2023defending} proposed the psychologically inspired self-reminder technique that can efficiently and effectively mitigate against jailbreaks without further training.
Unfortunately, research on the robustness of foundational and fine-tuned models is still limited.

\section{Responsible Open Source and Explanation}
Lack of transparency of models behind AIGC can lead to a series of unsatisfactory results. It is frequently challenging to  explain why and how the models generate different content and determine the information used to generate a model's output. 
For example, social and cultural bias is introduced and potentially amplified at multiple stages of model development and deployment. However, how the biases are propagated through these models remain unclear. Similarly, while deduplication can be an effective method of preventing memorization, it does not completely explain why or how models like DALL·E 2 memorize training data.
As most of the code and models behind AIGC are not transparent to the public, and their downstream applications are diverse and may have complex societal impacts, it is challenging to determine the potential harms they may cause. Therefore, the need for responsible open source becomes critical in determining whether the benefits of AIGC outweigh its potential risks in specific use cases. Open-sourcing can also facilitate 
explanation of the behaviours of the models behind AIGC.

Currently, most companies chose not to release their models or open-source their code before solving all the potential risks associated with their models. OpenAI has been criticized for not sharing more about how the most recent GPT-4 was created. Stable Diffusion~\cite{stable-diffusion-url} and Meta's LLAMA~\cite{llama-url} 
are 
few generative AI models that provide the source code and pretrained model (weights). The risk is that anyone can use 
these open-sourced models for free, even for commercial or malicious purposes. 
To promote a healthy open-sourcing environment, communities have put a lot of joint efforts. In Dec, 2023, IBM And Meta Launch the AI Alliance for safe and open AI with the belief that open and transparent innovation is crucial for harnessing AI advancements in a way that prioritizes safety, diversity, and widespread economic opportunity~\cite{forbes-ai-alliance}.


\section{Limit Technology Abuse}
AIGC can be used for malicious purposes such as spreading fake news, hoaxes, and harassment. The foundation models that power AIGC have made it easier and cheaper to create deepfakes that are close to the original, posing additional risks and concerns. In fact, many models are still far from satisfactory and some of them have gained negative reputations for producing useless, biased, or harmful information.

For example, on the 4chan online forum, there are numerous discussions about images of naked celebrities and other forms of fake pornographic content generated by Stable Diffusion~\cite{deepfake-url}. The misuse of these technologies could lead to the spread of misinformation, harm the reputations of individuals, or even break the law.
The potential negative impact of ChatGPT on education is significant, as students could use it to write homework or solve math problems, thus compromising the integrity of their work. Moreover, as ChatGPT is a chatbot, it lacks the necessary emotional connection that a human teacher can provide, which could lead to a diminished learning experience. In light of these concerns, New York City public schools have recently banned the use of ChatGPT~\cite{chatgpt-school-banned-url}.
Stack Overflow, a Q\&A platform for coders and programmers, temporarily prohibited the sharing of ChatGPT information, acknowledging its potential to cause significant harm to the site and users who rely on it for accurate answers~\cite{chatgpt-stackoverflow-banned-url}. Writing and editing tools that rely on ChatGPT also face the risk of losing customers if they inadvertently introduce errors into the output.

Overall, the potential misuse of AIGC poses a threat to not only the users but also the whole creative industry. Therefore, it is crucial to use AIGC only in situations where the risk can be managed or corrected. To mitigate risks, it is also necessary to include governance mechanisms for AIGC models as soon as possible, such as establishing legal regulations. The most recent deal on comprehensive rules for trustworthy AI from EU~\cite{EU-AI-Act} reflects the urgency to deal with concerns on the misuse of AIGC technologies.

\section{Consent, credit, and compensation}
Many AIGC models are trained on datasets without obtaining consent or providing credit or compensation to the original data contributors. 
For example, Simon Willison and Andy Baio found that a large number of images in LAION were copied from DeviantArt and used to train Stable Diffusion~\cite{exploring-the-training-data-url}. This results in data contributors' works being learned by AI models and recreated by other users for profit, without their knowledge or permission. This practice damages the interests of the original data contributors. To avoid negative impacts, AIGC companies should obtain consent from data contributors and take proactive measures before training their models on any original or augmented works. Failure to do so could result in lawsuits against AIGC. Therefore, AIGC companies must ensure that data collection and model training are conducted in an ethical and responsible manner.

A potential solution to the issue of using creators' works for AI training is to notify them from the beginning and give them the option to benefit from subsequent creations based on their works generated by the model. Additionally, creators who give their consent for their data to be used can be rewarded based on how their creations contribute to AIGC each time the tool is queried. By incentivizing creators, companies can encourage creators to contribute more and accelerate the development of AIGC. For example, a more user-friendly version of Copilot could allow voluntary participation or compensate coders for contributing to the training corpus~\cite{github-copilot-investigation-url}.

\section{Environment impact}
The massive size of AIGC models, which can have billions even trillions of parameters, results in high environmental costs for both model training and operation. 
For example, GPT-3 has 175 billion parameters and requires significant computing resources to train. 
Narayanan \etal~\cite{narayanan2021efficient} estimated that training GPT-3 with A100s would require 1,024 GPUs, 34 days, and cost 4.6 million dollars, with an expected energy consumption of 936 MWh~\cite{AI-harming-url}. This raises important questions about how to reduce the energy consumption and carbon emission of AIGC models.

The upcoming GPT-4, with even more parameters than its predecessor, is expected to leave a more significant carbon emission. Failing to take appropriate steps to mitigate the substantial energy costs of AIGC could lead to irreparable damage to our planet. It is crucial to address these concerns and explore sustainable alternatives. Communities have started to explore more slim alternatives with decent performance as much larger ones.

\section{Discussion}
\label{sec:discussion}

\textbf{What about commercial usage: a vicious competition? Will AIGC replace humans and become a roadblock to human creativity?} 
Many AIGC models are being utilized for art and graphic design commercially. For example, PromptBase~\cite{promptbase-url} is an early marketplace for DALL·E, Midjourney, Stable Diffusion \& GPT-3 prompts. 
Microsoft is using DALL-E 2 to power a generative art feature that will be available in Microsoft Edge. Microsoft and OpenAI are collaborating on ChatGPT-Powered Bing~\cite{ip-concerns-url}. Moreover, Microsoft had integrated OpenAI's ChatGPT into Word, PowerPoint, Outlook, and other applications to allow users to automatically generate text using simple prompts~\cite{ghost-writer-url}. While using the generated works for profit or commercial purposes is not recommended, there are no mandatory legal restrictions at this stage.

The use of AIGC has faced criticism from those who fear that it will replace human jobs. 
Insider has listed several jobs that could potentially be replaced by ChatGPT, including coders, data analysts, journalists, legal assistants, traders, accountants, etc~\cite{chatgpt-replace-url}. Some artists worry that the wide use of image generation tools such as Stable Diffusion could eventually make human artists, photographers, models, cinematographers, and actors commercially uncompetitive~\cite{this-artist-url}. For example, the images generated by Stable Diffusion can be sold on the market. This creates direct competition and poses a significant threat to creators, such as writers, artists, and programmers, who could suffer permanent damage to their businesses~\cite{stable-diffusion-litigation-url}. 
Since Stable Diffusion can produce an unlimited number of infringing images, this threat is even more significant.
However, David Holz, the founder of Midjourney, views artists as customers rather than competitors. Artists can use Midjourney to quickly prototype artistic concepts and show them to clients before starting work themselves~\cite{david-holz-url}.

As AIGC models become more widespread, people may become too dependent on instant answers and less willing to think critically on their own, which could ultimately diminish or destroy human creativity and increase the risk of AI exerting control over humans. Overreliance on AIGC could create opportunities for malicious attackers to exploit user trust and access their private information.





\textbf{Fairness of benefit distribution.}
It is important to recognize that AIGC models may have varying impacts on different groups of people depending on their environmental and individual abilities, which could further exacerbate global inequities~\cite{weidinger2021ethical}. Addressing the issue of how to fairly distribute the benefits of AIGC models is an area that requires further exploration and attention.

\textbf{Conflict among multiple goals}.
It is critical to ensure that the mitigation of one risk does not exacerbate another~\cite{weidinger2021ethical}. For example, approaches to mitigate the use of toxic language in language models can introduce biases in model predictions against marginalized communities~\cite{welbl2021challenges,xu2021detoxifying}. Therefore, it is essential to explore effective mitigation strategies that can 
trade-off multiple goals.

\section{Conclusion}
Although AIGC is still in its infancy, it is rapidly expanding and will remain active for the foreseeable future. 
Current AIGC technologies only scratch the surface of what AI can create in the field of creativity. While AIGC offers many opportunities, it also carries significant risks. To acquire a thorough comprehension of these risks, we provide a synopsis of both current and potential threats in recent AIGC models, so that both the users and companies can be well aware of these risks, and make the appropriate actions to mitigate them.

In order to promote responsible usage of AIGC tools and mitigate associated risks, we propose several steps that companies and users can take. It is important for companies to incorporate responsible AI practices throughout 
the whole life cycles during development of AIGC products. For example, proactive measures should be taken to mitigate potential risks in data sources, models, and pre/post-processing steps. Without proper safeguards, AIGC development may face significant challenges and regulatory hurdles. Note that this vision paper is not exhaustive, and it is essential for the wider community to contribute to the understanding and implementation of responsible AIGC. To facilitate this, it is necessary to build comprehensive benchmarks for measuring and evaluating the risks associated with different AIGC technologies.

\bibliographystyle{named}
\bibliography{survey_biblio}
\end{sloppypar} 
\end{document}